\documentclass[letterpaper, 10 pt, conference]{ieeeconf}  

\IEEEoverridecommandlockouts                              

\usepackage{graphicx}
\usepackage{amssymb}
\usepackage{array}
\usepackage{multirow}
\let\labelindent\relax
\usepackage{enumitem}
\usepackage[font=small]{caption}
\usepackage[colorlinks]{hyperref}
\usepackage{dblfloatfix}
\usepackage{todonotes}

\newcolumntype{C}[1]{>{\centering\let\newline\\\arraybackslash\hspace{0pt}}m{#1}}
\newcolumntype{L}[1]{>{\raggedright\let\newline\\\arraybackslash\hspace{0pt}}m{#1}}
\newcolumntype{R}[1]{>{\raggedleft\let\newline\\\arraybackslash\hspace{0pt}}m{#1}}

\setlist[enumerate]{wide=0pt,leftmargin=*}
\setlist[itemize]{wide=0pt,leftmargin=*}

\title{\LARGE \bf
FewSOL: A Dataset for Few-Shot Object Learning in Robotic Environments
}

\author{
Jishnu Jaykumar P, Yu-Wei Chao, Yu Xiang
\thanks{Jishnu Jaykumar P and Yu Xiang are with the Department of Computer Science, University of Texas at Dallas, Richardson, TX 75080, USA \tt\small \{jishnu.p,yu.xiang\}@utdallas.edu}
\thanks{Yu-Wei Chao is with NVIDIA, Seattle, WA 98105, USA \tt\small ychao@nvidia.com}
}

\begin{document}

\maketitle
\thispagestyle{empty}
\pagestyle{empty}

\begin{abstract}

We introduce the Few-Shot Object Learning (\textsc{FewSOL}) dataset for object recognition with a few images per object. We captured 336 real-world objects with 9 RGB-D images per object from different views. \textsc{FewSOL} has object segmentation masks, poses, and attributes. In addition, synthetic images generated using 330 3D object models are used to augment the dataset. We investigated (i) few-shot object classification and (ii) joint object segmentation and few-shot classification with state-of-the-art methods for few-shot learning and meta-learning using our dataset. The evaluation results show the presence of a large margin to be improved for few-shot object classification in robotic environments, and our dataset can be used to study and enhance few-shot object recognition for robot perception\footnote{Dataset and code available at \\ \url{https://irvlutd.github.io/FewSOL}}. 
\end{abstract}

\section{Introduction}
	
For robots to work in human environments, they will encounter various objects in our daily lives. How can we build models to enable robots to recognize all kinds of objects and eventually manipulate these objects? In the robotics community, model-based object recognition has been the focus, where 3D models of objects are built and used for recognition. For example, the YCB Object and Model Set~\cite{calli2015benchmarking} has significantly benefited 6D object pose estimation and manipulation research. The limitation of model-based object recognition is that it is difficult to obtain a large number of 3D models for many objects in the real world. The 3D scanning techniques are expensive and certain object categories such as reflective objects and transparent objects cannot be reconstructed well. Another paradigm for object recognition focuses on recognizing object categories such as bowls, mugs and bottles. Most datasets for object category recognition only contain a few dozen categories. For instance, the MSCOCO dataset for object detection and segmentation~\cite{lin2014microsoft} has 80 categories. The NOCS dataset for object category pose estimation~\cite{wang2019normalized} only has 6 categories. While large-scale datasets collected from the Internet such as ImageNet~\cite{deng2009imagenet}, Visual Genome~\cite{krishna2017visual}, Objects365~\cite{shao2019objects365}, and open images~\cite{kuznetsova2020open} contain large numbers of object categories, these datasets are useful for learning visual representations but are not very suitable to learn object representations for robot manipulation due to the domain differences.




Few-shot learning~\cite{wang2020generalizing} emphasizes learning from a few examples per object, which has the potential to overcome the limitations of model-based and category-based approaches. However, most datasets for few-shot learning in the literature focus on image classification using images from the Internet. In this work, we introduce a new dataset to facilitate few-shot object recognition in robotic environments. Our aspiration is that if robots can recognize objects from a few exemplar images, it is possible to scale up the number of objects a robot can recognize since collecting a few images per object is a much easier process compared to building a 3D model of an object. In addition, models trained in the meta-learning setting~\cite{finn2017model} can generalize to new objects without re-training.

\begin{figure*}
	\centering
	\includegraphics[width=0.8\textwidth]{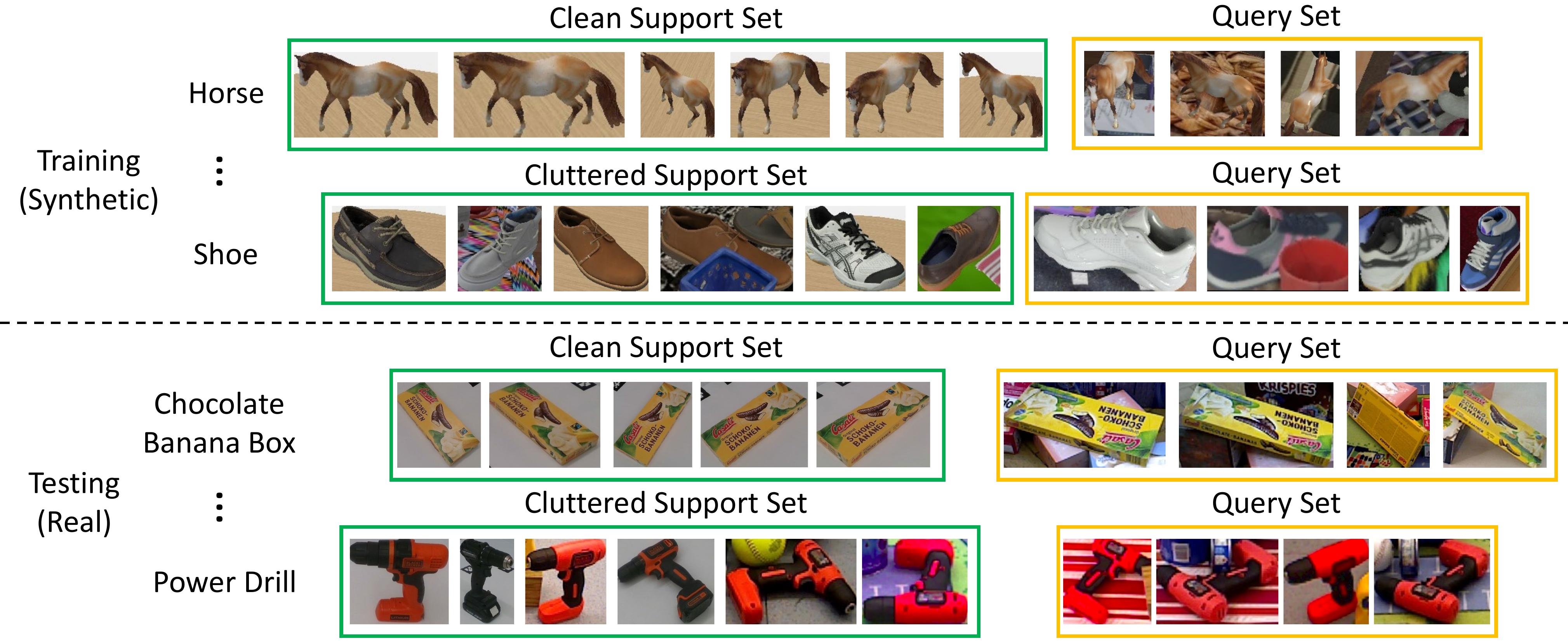}
	\caption{Examples of support sets and query sets from our dataset. Clean support sets only contain images with single objects in the clean background, while cluttered support sets have images with multiple objects and different backgrounds.}
	\label{fig:support_query}
	\vspace{-2mm}
\end{figure*}

In our Few-Shot Object Learning (\textsc{FewSOL}) dataset, we have collected images for 336 objects in the real world. For each object, we collected 9 RGB-D images from different views, i.e., 9 shots per object. We provide ground truth segmentation masks and 6D object poses of these objects, where the object poses are computed using AR tags. In addition, we employed Amazon Mechanical Turk (MTurk) to collect annotations of these objects including object names, object categories, materials, function and colors. For each object, we collected annotations from 5 MTurkers and then merged their answers on the object attributes. This way, we can account for how different people name these objects in our dataset. Based on the collected object names, we have defined 198 classes for these 336 objects. In few-shot learning or meta-learning settings, we can think of these images as support sets. Our goal is to apply learned models to cluttered scenes. Therefore, we include the images from the Object Clutter Indoor Dataset (OCID)~\cite{suchi2019easylabel} in our dataset. Segmentation masks of objects are provided in the OCID dataset. We manually annotated the class names of these objects and found that they belong to 52 classes in our object categories. These images can be used as query sets.

\begin{table*}[t]
\centering
\resizebox{0.9\linewidth}{!}{\begin{tabular}{|cccccc|}
\hline
Dataset  & Class type  & \#classes & Image Type & \#images\_per\_class & Annotations \\ \hline
Omniglot~\cite{lake2015human} & Characters & 1623 &  RGB & 20 & class label \\
mini-ImageNet~\cite{vinyals2016matching} & WordNet synsets & 100 & RGB & 600 & class label\\
ILSVRC-2012~\cite{russakovsky2015imagenet} & WordNet synsets & 1000 & RGB & $\approx$8,004 & class label\\
Aircraft~\cite{maji2013fine} & Aircraft & 100 & RGB & 100 & class label\\
CUB-200-2011~\cite{wah2011caltech} & Birds & 200 & RGB & $\approx$59 & class label\\
Describable Texture~\cite{cimpoi2014describing} & Textures & 47 & RGB & 120 & class label\\
Quick Draw~\cite{Jongejan2016quick} & Drawings & 345 & RGB & $\approx$146,164 & class label\\
Fungi~\cite{sulc2020fungi} & Fungal species & 1394 & RGB & $\approx$65 & class label\\
VGG Flower~\cite{nilsback2008automated} & Flowers & 102 & RGB & $\approx$81 & class label\\
Traffic Signs~\cite{houben2013detection} & Traffic signs & 43 & RGB & $\approx$912 & class label\\
MSCOCO~\cite{lin2014microsoft} & Internet Objects & 80 & RGB & $\approx$10,751 & class label, segmentation\\
\textbf{Ours (real + synthetic)} & \textbf{Daily objects} & \textbf{198 + 125} & \textbf{RGB-D} & \textbf{$\approx$27 + 10,234}
 & \textbf{class label, segmentation, object pose and attribute} \\ \hline
\end{tabular}}
\caption{Comparison of our dataset with other datasets for few-shot learning in the literature. Our dataset contains daily objects in robot manipulation settings with both real and synthetic images and additional annotations other than object class label.}
\label{table:dataset}
\vspace{-6mm}
\end{table*}

To further expand the scale of our dataset, we also generate synthetic data to complement the data from the real world. We selected 330 3D object models from the Google Scanned Objects dataset~\cite{downs2022google} and used the PyBullet simulator to compose synthetic scenes of these objects and render synthetic RGB-D images from the scenes. Similar to the real-world data, we first put each 3D model onto a table and generate 9 views. The benefit of using synthetic data is that we can generate cluttered scenes with these objects and obtain annotations for all the objects. We generate 40,000 cluttered scenes and render 7 views per scene.

In this paper, we use our dataset to study two problems: (i) few-shot object classification and (ii) joint object segmentation and few-shot classification. For few-shot classification, we follow the protocol proposed in the Meta-Dataset~\cite{triantafillou2019meta} which constructs episodes for training and testing. Each episode consists of multiple support sets and query sets with different number of classes and images per class. Fig.~\ref{fig:support_query} illustrates some support sets and query sets from our dataset. We reserve the cluttered images from the OCID dataset for testing and limit training to synthetic data only. In this way, we can investigate sim-to-real transfer for few-shot learning with our dataset. We use the ground truth segmentation masks to crop the objects for classification. For joint object segmentation and few-shot classification, we first apply an object segmentation method and then use the predicted masks to crop the objects for classification. Therefore, segmentation errors can be accounted for in the classification accuracy. We have evaluated state-of-the-art methods for few-shot learning and meta-learning in these two settings using our dataset. These results can be used for future comparisons. To the best of our knowledge, our dataset is the first large-scale dataset for few-shot object learning. Enabling robots to recognize these object categories in our dataset can provide information of objects to downstream tasks such as manipulation, object retrieval or human-robot interaction using object names.



\begin{figure}
	\centering
	\includegraphics[width=.42\textwidth]{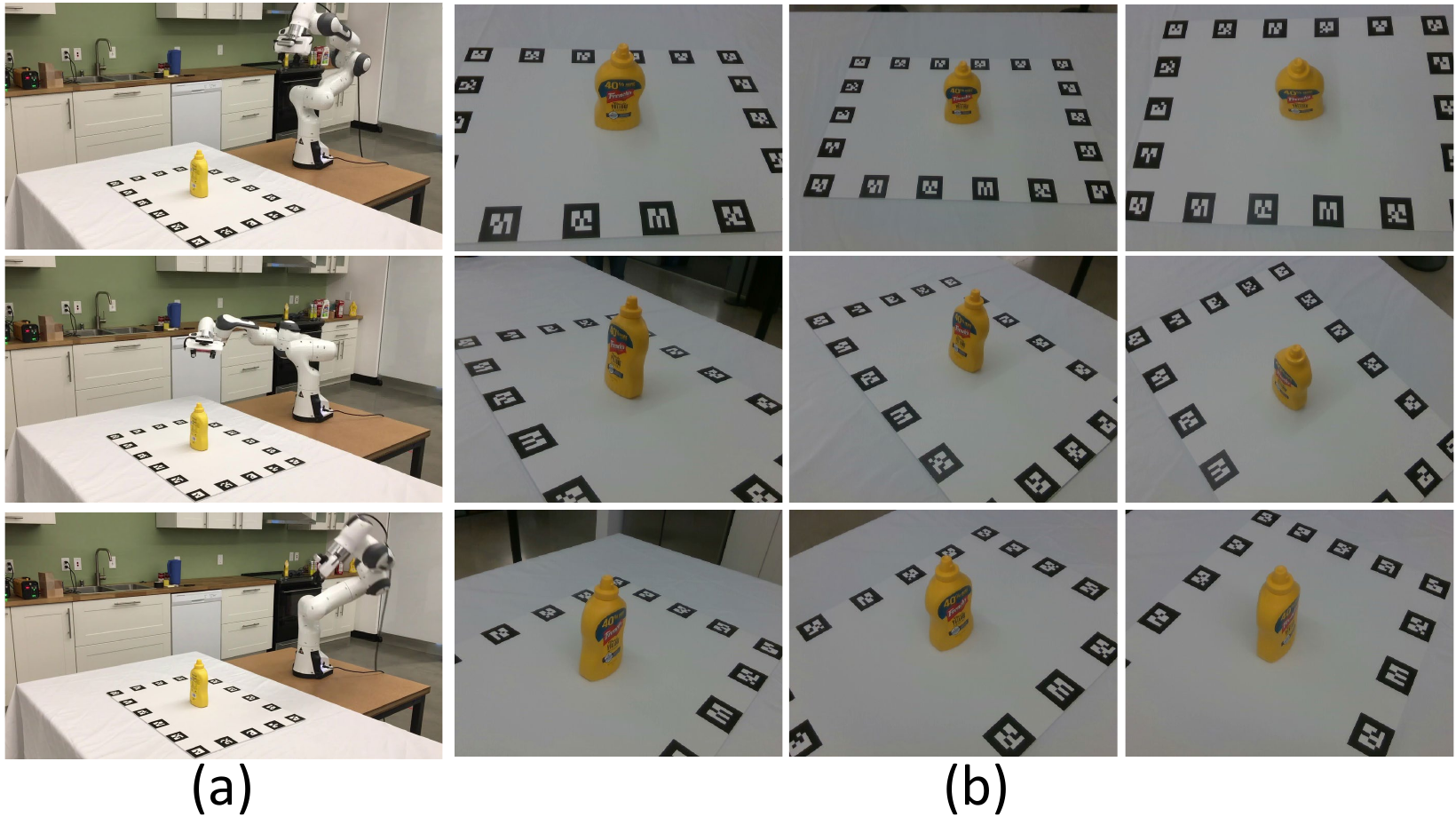}
	\caption{(a) Our data capture system with a Franka Emika Panda arm. (b) 9 images of a mustard bottle captured from different views.}
	\label{fig:data_capture}
	\vspace{-8mm}
\end{figure}

\begin{figure*}
	\centering
	\includegraphics[width=0.9\textwidth]{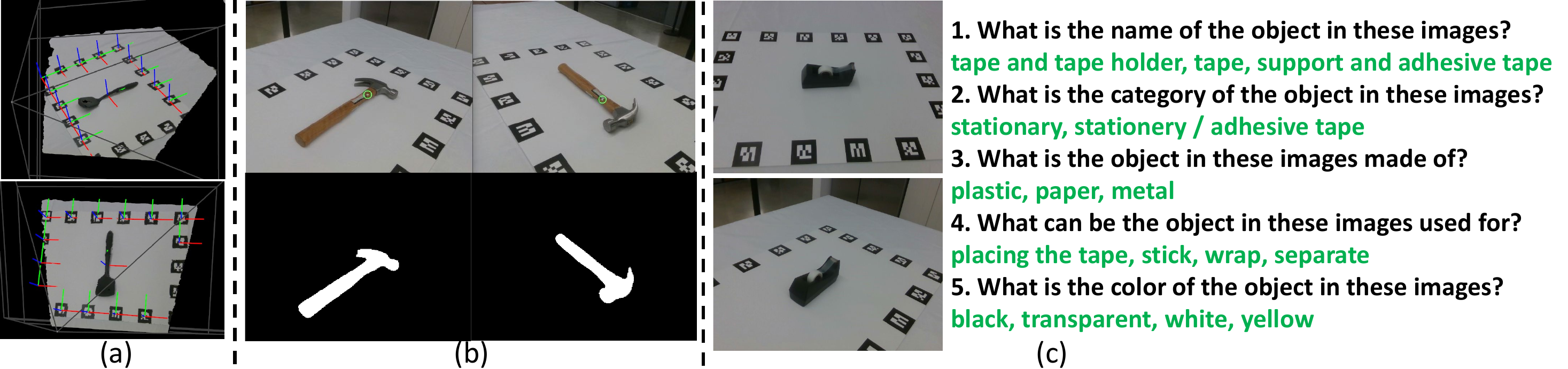}
	\caption{(a) Object poses from AR tags (b) Pixel correspondences using computed object poses and the segmentation masks of the objects. (c) Our Amazon Mechanical Turk questionnaire for object annotation.}
	\label{fig:camera_pose}
	\vspace{-4mm}
\end{figure*}

\section{Related Work}
\label{sec:citations}

\textbf{Few-Shot Learning and Meta-Learning.} In the context of image classification, few-shot learning indicates using a few images per class. The problem is usually formulated as ``$N$-way, $k$-shot'', i.e., $N$ classes with $k$ images per class. The end goal of few-shot learning is to learn a model on a set of training classes $\mathcal{C}_{train}$ that can generalize to novel classes $\mathcal{C}_{test}$ in testing. Each class has a support set and a query set. While the ground truth labels of both the support set and the query set for a class in $\mathcal{C}_{train}$ are available for learning, for a testing class in $\mathcal{C}_{test}$, only labels of the support set are available. Non-episodic approaches using all the data in $\mathcal{C}_{train}$ for training such as $k$-NN and its `Finetuned' variants~\cite{gidaris2018dynamic,qi2018low,chen2019closer,tian2020rethinking}. These methods focus on learning feature representations using neural networks that can be used in $\mathcal{C}_{test}$. Episodic approaches are considered to be meta-learners. An episode in training or testing consists of a subset of classes with support and query sets. Learning is performed by minimizing the loss on the query sets of the training episode. Representative episodic approaches include Prototypical Networks~\cite{snell2017prototypical}, Matching Networks~\cite{vinyals2016matching}, Relation Networks~\cite{sung2018learning}, Model Agnostic Meta-Learning (MAML)~\cite{finn2017model}, Proto-MAML~\cite{triantafillou2019meta} and CrossTransformers~\cite{doersch2020crosstransformers}. We evaluated majority of these state-of-the-art few-shot learning methods on \textsc{FewSOL}.

\textbf{Datasets for Few-Shot Learning.} The Omniglot~\cite{lake2015human} and the mini-ImageNet~\cite{vinyals2016matching} are widely used for evaluating few-shot learning methods in the literature. Recently, the Meta-Dataset~\cite{triantafillou2019meta} was introduced for benchmarking few-shot learning and meta-learning methods. Meta-Dataset leverages data from 10 different datasets: ILSVRC-2012 (ImageNet~\cite{russakovsky2015imagenet}), Omniglot~\cite{lake2015human}, Aircraft~\cite{maji2013fine}, CUB-200-2011 (Birds~\cite{wah2011caltech}), Describable Textures~\cite{cimpoi2014describing},
Quick Draw~\cite{Jongejan2016quick}, Fungi~\cite{sulc2020fungi}, VGG Flower~\cite{nilsback2008automated}, Traffic Signs~\cite{houben2013detection} and MSCOCO~\cite{lin2014microsoft}. As we can see, these data do not include daily objects for robot manipulation. Our dataset complements existing datasets for few-shot learning by explicitly addressing robotic applications. Table~\ref{table:dataset} compares our FewSOL dataset with existing datasets for few-shot learning. We provide both real and synthetic RGB-D images and more annotations for objects. It is worth mentioning that the CO3D~\cite{reizenstein2021common} dataset for 3D reconstruction has the potential to be used for few-shot object learning.


\vspace{-2mm}
\section{Dataset Construction}



\subsection{Data Capture in the Real World}

For each object in the real world, we capture multiple exemplar images of the object. We automate this process using a Franka Emika Panda arm as shown in Fig.~\ref{fig:data_capture}(a). An Intel RealSense D415 camera is mounted onto the Panda arm gripper. It can capture both RGB images and depth images. We specify 9 waypoints of the camera pose and utilize motion planning of the arm to move the camera to these poses. Due to the kinematic constraints of the robot arm, we cannot capture images for the backside of an object. For each object, we can automatically capture 9 RGB-D images from 9 different views (Fig.~\ref{fig:data_capture}(b), 3 poses on the left, 3 poses in the front and 3 poses on the right). 

To accurately estimate the camera poses of these 9 images with respect to the object, we have designed a marker board with 18 AR tags. During data capture, we place the object in the center of the board. For each captured image, we use the detected AR tag poses to compute the camera pose with respect to the center of the marker board and treat this pose as the estimated object pose of the image. Because there are noises in the AR-tag poses, we use the RANSAC algorithm to estimate the object pose in this process. Fig.~\ref{fig:camera_pose}(a) shows the AR-tag poses and the computed object poses with the point clouds from the RealSense camera. Using the object poses and the point clouds, we can also compute pixel correspondences between images as shown in Fig.~\ref{fig:camera_pose}(b). Consequently, we obtain 9 RGB-D images with their estimated object poses for each object. We have captured 336 objects in total. These include various daily objects from grocery stores and tools.

\subsection{Data Annotation} \label{sec:annotation}

After capturing these objects, our next step is to provide annotations. First, we generate the segmentation masks of these captured objects. Instead of manually segmenting these objects, we utilize the unseen object instance segmentation method proposed in~\cite{xiang2021learning}. The trained network in~\cite{xiang2021learning} cannot successfully segment all the objects in the beginning. Therefore, we bootstrap the network by finetuning it on our dataset. After applying the network to all the data, we manually select accurate segmentation for finetuning and then apply the finetuned network to unsuccessful images. We iterate this process until the network can segment all the objects. Fig.~\ref{fig:camera_pose}(b) shows two examples of the generated segmentation masks.


Second, we provide object class labels and additional attributes for these objects. We leverage Amazon Mechanical Turk (MTurk) for this annotation. In this way, we can gather how lay people name the classes and attributes of the objects, which can be useful to deploy object recognition systems to human-robot interaction scenarios where users communicate with robots using these common names. We designed 5 questions for each object and ask MTurkers to answer these questions. For each object, we gather answers from 5 different MTurkers and merge their answers. Fig.~\ref{fig:camera_pose}(c) illustrates an example with the questions and the merged answers. These annotations can be used to recognize detailed attributes of objects. Based on these annotations from MTurk, we define 198 object classes for these 336 captured objects. Since 9 images are captured for each object, each class has around 15 images in our dataset on average.

\subsection{Synthetic Data Generation}

\begin{figure}[t]
	\centering
	\includegraphics[width=0.4\textwidth]{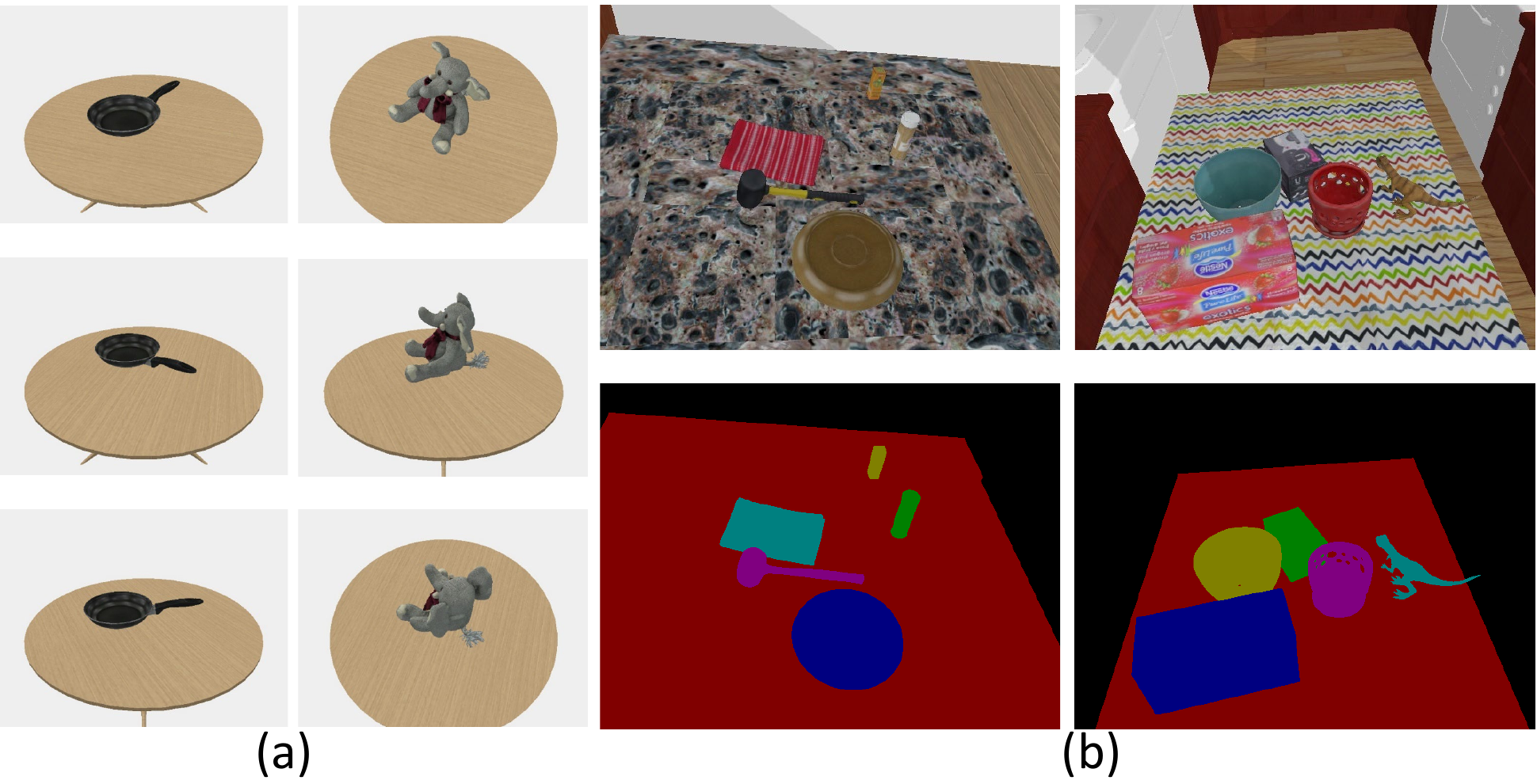}
	\caption{(a) Synthetic objects with clean background. (b) Synthetic objects in cluttered scenes.}
	\label{fig:synthetic}
	\vspace{-4mm}
\end{figure}

Leveraging synthetic data for learning has been successful in various robotic problems such as object segmentation~\cite{xiang2021learning}, grasping~\cite{mahler2017dex} and control policy learning~\cite{chebotar2019closing} since one can generate large-scale synthetic data with ground truth annotations automatically. In our dataset, we also leverage synthetic images for few-shot object learning. To do so, we selected 330 3D object models from Google Scanned Objects~\cite{downs2022google}. We use these 3D models to generate two types of data. First, similar to our real-world data capture, we place each object onto a table in the PyBullet simulator and generate 9 RGB-D images of each object from 9 different views. Fig~\ref{fig:synthetic}(a) shows some examples of these multi-view images. Second, we generate cluttered scenes using these objects as shown in Fig~\ref{fig:synthetic}(b). Thanks to the simulator, we can obtain object segmentation masks of these images effortlessly, which is otherwise time-consuming to collect from the cluttered scenes in the real world. We generated 40000 scenes of these objects on tabletops and rendered 7 RGB-D images per scene. To obtain class names of these 330 objects, we also employ MTurk to collect annotations of these objects as described in Sec.~\ref{sec:annotation}. Eventually, we define 125 classes for these synthetic objects.


\begin{figure}[t]
  \centering 
 \includegraphics[width=0.4\textwidth ]{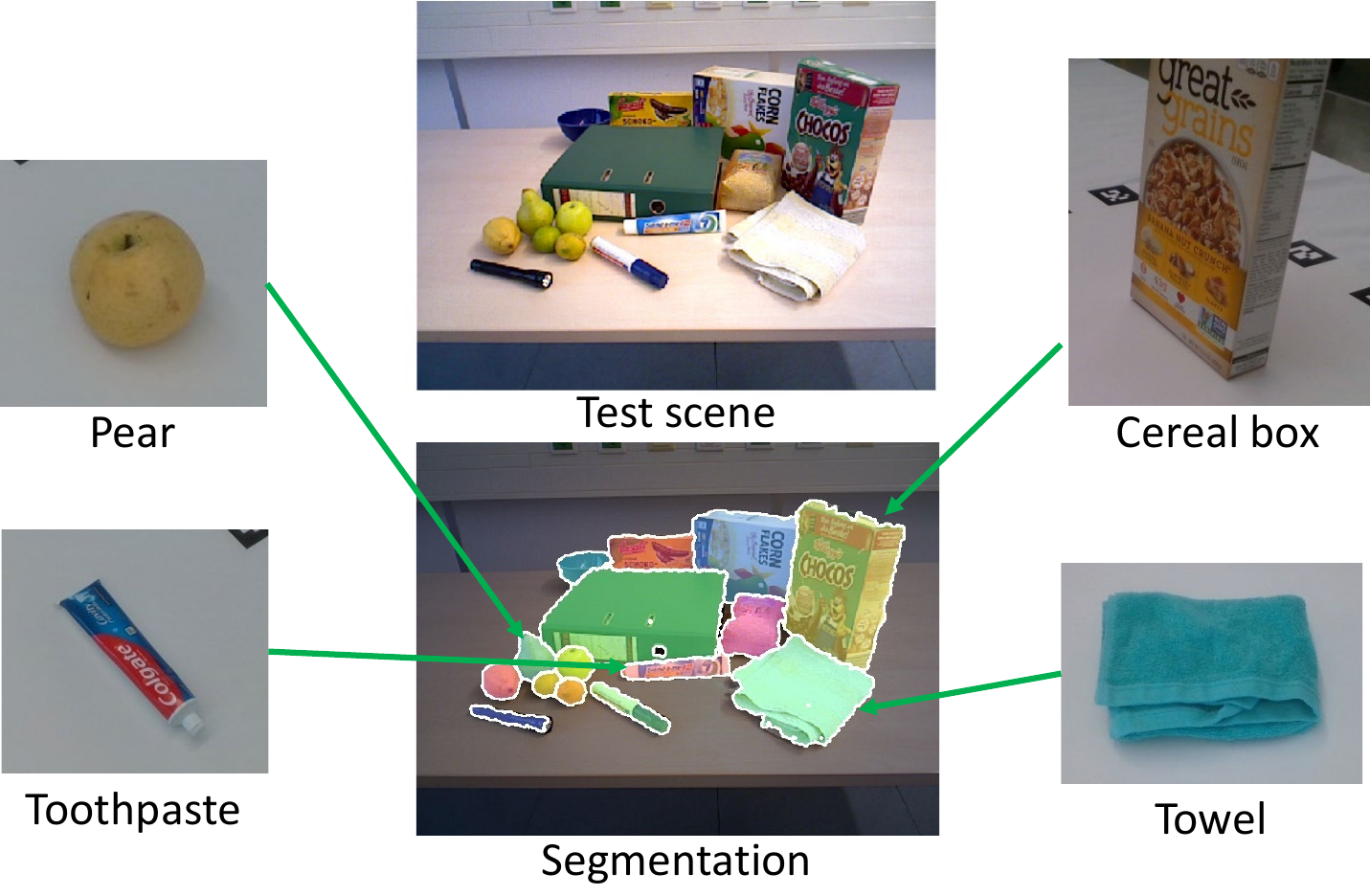}
   \caption{Illustration of the joint object segmentation and few-shot classification problem with an image from the OCID dataset~\cite{suchi2019easylabel}.}
   \label{fig:segmentation}
    \vspace{-5mm}
\end{figure}

\subsection{Joint Object Segmentation and Few-Shot Classification}

In real-world robotic applications, objects usually appear in cluttered scenes. Therefore, we need to separate an object from other objects plus the background and then classify it. For our dataset, we target the problem of joint object segmentation and few-shot classification. The problem is illustrated in Fig.~\ref{fig:segmentation}. Given an image of a cluttered scene, the task is to segment objects in the scene and classify each object in few-shot learning settings. To evaluate the performance of joint object segmentation and few-shot classification on real-world images, we leverage the Object Clutter Indoor Dataset (OCID) proposed in~\cite{suchi2019easylabel}. This dataset was originally proposed for object segmentation. It provides segmentation masks of objects in the dataset. We manually annotate objects in the OCID dataset with class labels. We found 2,300 objects in the OCID dataset after filtering a few bad segmentation annotations. These objects belong to 52 classes in our dataset. Objects in OCID can be partially occluded which makes the few-shot object classification challenging.

\begin{table*}[t]
\centering
\resizebox{0.75\linewidth}{!}{\begin{tabular}{|c|cc|cc|cc|cc|}
\hline
\multirow{3}{*}{Method} &   \multicolumn{6}{c|}{OCID (Real)~\cite{suchi2019easylabel}} & \multicolumn{2}{c|}{Google (Synthetic)~\cite{downs2022google}} \\ \cline{2-9}
& \multicolumn{2}{c|}{All (52 classes)} & \multicolumn{2}{c|}{Unseen (41 classes)} & \multicolumn{2}{c|}{Seen (11 classes)} & \multicolumn{2}{c|}{Unseen (13 classes)} \\ & Cluttered S & Clean S & Cluttered S & Clean S & Cluttered S & Clean S & Cluttered S & Clean S \\ \hline
\multicolumn{9}{|c|}{Training setting: clean support set without pre-training} \\ \hline
$k$-NN~\cite{triantafillou2019meta} & 52.92 $\pm$ 1.08 & 16.19 $\pm$ 0.74 & 55.12 $\pm$ 1.08 & 16.73 $\pm$ 0.62 & 55.90 $\pm$ 1.08 & 31.94 $\pm$ 0.89 & 83.43 $\pm$ 0.76 & 79.91 $\pm$ 0.79 \\
Finetune~\cite{triantafillou2019meta} & \textbf{57.96 $\pm$ 1.08} & 28.99 $\pm$ 0.92 & \textbf{62.45 $\pm$ 1.13} & \textbf{33.14 $\pm$ 0.91} & 58.49 $\pm$ 1.01 & 36.46 $\pm$ 1.04 & 63.36 $\pm$ 1.75 & 66.60 $\pm$ 1.19 \\
ProtoNet~\cite{snell2017prototypical} & 45.02 $\pm$ 0.89 & 16.18 $\pm$ 0.74 & 50.00 $\pm$ 0.91 & 17.20 $\pm$ 0.72 & 48.90 $\pm$ 0.88 & 32.73 $\pm$ 0.89 & 71.62 $\pm$ 0.98 & 72.63 $\pm$ 0.79 \\
MatchingNet~\cite{vinyals2016matching} & 51.58 $\pm$ 1.08 & 19.40 $\pm$ 0.72 & 58.24 $\pm$ 1.04 & 21.70 $\pm$ 0.76 & 53.99 $\pm$ 1.02 & 33.10 $\pm$ 0.94 & 76.26 $\pm$ 0.82 & 77.26 $\pm$ 0.76 \\
fo-MAML~\cite{finn2017model} & 24.24 $\pm$ 1.17 & 17.01 $\pm$ 1.00 & 30.29 $\pm$ 1.33 & 19.09 $\pm$ 0.86 & 41.82 $\pm$ 0.95 & 41.82 $\pm$ 0.95 & 51.31 $\pm$ 1.70 & 59.54 $\pm$ 0.96 \\
fo-Proto-MAML~\cite{triantafillou2019meta} & 49.57 $\pm$ 1.00 & 20.06 $\pm$ 0.79 & 55.96 $\pm$ 1.04 & 22.51 $\pm$ 0.73 & 55.32 $\pm$ 1.03 & 33.45 $\pm$ 0.95 & 68.70 $\pm$ 1.42 & 81.69 $\pm$ 0.80 \\
CTX~\cite{doersch2020crosstransformers} & 53.17 $\pm$ 1.05 & 18.49 $\pm$ 0.84 & 55.81 $\pm$ 0.99 & 20.60 $\pm$ 0.94 & 56.77 $\pm$ 0.98 & 35.41 $\pm$ 0.93 & \textbf{86.46 $\pm$ 0.70} & \textbf{88.08 $\pm$ 0.63} \\
CTX+SimCLR~\cite{doersch2020crosstransformers} & 53.87 $\pm$ 1.03 & \textbf{30.31 $\pm$ 1.00} & 56.56 $\pm$ 0.97 & 30.43 $\pm$ 0.93 & \textbf{64.90 $\pm$ 0.98} & \textbf{53.70 $\pm$ 1.18} & 85.15 $\pm$ 0.69 & 83.94 $\pm$ 0.65 \\ \hline
\multicolumn{9}{|c|}{Training setting: cluttered support set without pre-training} \\ \hline
$k$-NN~\cite{triantafillou2019meta} & 58.78 $\pm$ 1.04 & 21.72 $\pm$ 0.84 & 61.56 $\pm$ 1.13 & 22.17 $\pm$ 0.78 & \textbf{66.33 $\pm$ 1.02} & 41.76 $\pm$ 1.00 & 86.94 $\pm$ 0.67 & 80.99 $\pm$ 0.69 \\
Finetune~\cite{triantafillou2019meta} & 58.63 $\pm$ 1.12 & 29.94 $\pm$ 0.90 & 63.18 $\pm$ 1.06 & 32.71 $\pm$ 0.89 & 59.28 $\pm$ 1.07 & 39.71 $\pm$ 1.01 & 66.36 $\pm$ 1.77 & 65.02 $\pm$ 1.24 \\
ProtoNet~\cite{snell2017prototypical} & 42.56 $\pm$ 0.88 & 15.17 $\pm$ 0.82 & 47.60 $\pm$ 0.87 & 16.23 $\pm$ 0.77 & 48.93 $\pm$ 0.89 & 33.69 $\pm$ 1.02 & 71.21 $\pm$ 0.97 & 66.76 $\pm$ 0.87 \\
MatchingNet~\cite{vinyals2016matching} & 52.94 $\pm$ 1.09 & 17.98 $\pm$ 0.77 & 56.20 $\pm$ 1.05 & 19.52 $\pm$ 0.73 & 54.07 $\pm$ 1.03 & 31.18 $\pm$ 0.93 & 78.51 $\pm$ 0.82 & 72.25 $\pm$ 0.87 \\
fo-MAML~\cite{finn2017model} & 43.92 $\pm$ 1.07 & 17.26 $\pm$ 0.87 & 49.21 $\pm$ 1.03 & 18.80 $\pm$ 0.80 & 51.94 $\pm$ 1.00 & 28.91 $\pm$ 0.92 & 70.78 $\pm$ 0.90 & 66.54 $\pm$ 0.88 \\
fo-Proto-MAML~\cite{triantafillou2019meta} & 51.00 $\pm$ 1.02 & 17.35 $\pm$ 0.75 & 55.46 $\pm$ 1.06 & 19.59 $\pm$ 0.74 & 56.90 $\pm$ 1.06 & 31.99 $\pm$ 0.91 & 76.78 $\pm$ 1.10 & 77.36 $\pm$ 0.83 \\
CTX~\cite{doersch2020crosstransformers} & 49.96 $\pm$ 1.04 & 18.43 $\pm$ 0.74 & 53.91 $\pm$ 1.02 & 20.82 $\pm$ 0.87 & 57.97 $\pm$ 0.98 & 36.93 $\pm$ 1.03 & \textbf{92.45 $\pm$ 0.46} & 89.82 $\pm$ 0.58 \\
CTX+SimCLR~\cite{doersch2020crosstransformers} & \textbf{60.83 $\pm$ 1.06} & \textbf{31.67 $\pm$ 0.97} & \textbf{63.80 $\pm$ 1.09} & \textbf{33.34 $\pm$ 0.99} & 66.25 $\pm$ 1.01 & \textbf{51.62 $\pm$ 1.10} & 89.58 $\pm$ 0.57 & \textbf{88.99 $\pm$ 0.57} \\ \hline
\multicolumn{9}{|c|}{Training setting: clean support set with pre-training} \\ \hline
$k$-NN~\cite{triantafillou2019meta} & 59.34 $\pm$ 1.10 & 23.40 $\pm$ 0.85 & 63.02 $\pm$ 1.07 & 24.98 $\pm$ 0.86 & 66.24 $\pm$ 1.01 & 39.36 $\pm$ 1.01 & 89.18 $\pm$ 0.59 & 85.77 $\pm$ 0.69 \\
Finetune~\cite{triantafillou2019meta} & \textbf{59.77 $\pm$ 1.08} & 32.15 $\pm$ 0.90 & \textbf{64.01 $\pm$ 1.08} & 35.54 $\pm$ 0.90 & 58.30 $\pm$ 1.09 & 37.72 $\pm$ 1.04 & 66.09 $\pm$ 1.72 & 69.85 $\pm$ 1.09 \\
ProtoNet~\cite{snell2017prototypical} & 57.54 $\pm$ 1.06 & \textbf{34.47 $\pm$ 1.00} & 61.25 $\pm$ 1.11 & \textbf{37.06 $\pm$ 1.01} & 65.90 $\pm$ 1.04 & 51.07 $\pm$ 1.04 & 74.37 $\pm$ 1.12 & 81.38 $\pm$ 0.67 \\
MatchingNet~\cite{vinyals2016matching} & 53.81 $\pm$ 1.02 & 26.33 $\pm$ 0.94 & 57.77 $\pm$ 0.93 & 28.05 $\pm$ 1.00 & 61.83 $\pm$ 0.97 & 45.81 $\pm$ 1.07 & 65.40 $\pm$ 1.57 & 85.18 $\pm$ 0.70 \\
fo-MAML~\cite{finn2017model} & 44.92 $\pm$ 1.20 & 15.71 $\pm$ 0.77 & 51.67 $\pm$ 1.13 & 17.74 $\pm$ 0.77 & 56.02 $\pm$ 1.04 & 30.78 $\pm$ 0.95 & 70.91 $\pm$ 1.08 & 73.86 $\pm$ 0.88 \\
fo-Proto-MAML~\cite{triantafillou2019meta} & 57.09 $\pm$ 1.04 & 27.01 $\pm$ 0.94 & 60.29 $\pm$ 1.02 & 28.69 $\pm$ 0.88 & 66.75 $\pm$ 1.04 & 44.39 $\pm$ 1.10 & 77.16 $\pm$ 1.10 & 88.14 $\pm$ 0.68 \\
CTX~\cite{doersch2020crosstransformers} & 56.65 $\pm$ 1.02 & 29.06 $\pm$ 0.99 & 60.33 $\pm$ 1.02 & 29.96 $\pm$ 0.94 & 65.47 $\pm$ 1.04 & 45.48 $\pm$ 1.12 & \textbf{90.66 $\pm$ 0.63} & \textbf{92.72 $\pm$ 0.45} \\
CTX+SimCLR~\cite{doersch2020crosstransformers} & 57.47 $\pm$ 1.03 & 31.29 $\pm$ 0.98 & 59.32 $\pm$ 0.98 & 31.31 $\pm$ 0.93 & \textbf{67.73 $\pm$ 0.91} & \textbf{53.67 $\pm$ 1.14} & 81.76 $\pm$ 0.74 & 82.76 $\pm$ 0.74 \\ \hline
\multicolumn{9}{|c|}{Training setting: cluttered support set with pre-training} \\ \hline
$k$-NN~\cite{triantafillou2019meta} & 60.63 $\pm$ 1.10 & 23.09 $\pm$ 0.87 & 62.54 $\pm$ 1.12 & 23.97 $\pm$ 0.77 & 65.16 $\pm$ 1.04 & 40.82 $\pm$ 1.04 & 89.21 $\pm$ 0.55 & 84.49 $\pm$ 0.74 \\
Finetune~\cite{triantafillou2019meta} & 60.11 $\pm$ 1.12 & 31.23 $\pm$ 0.95 & 62.58 $\pm$ 1.10 & 33.22 $\pm$ 0.94 & 58.89 $\pm$ 1.03 & 36.48 $\pm$ 1.06 & 66.49 $\pm$ 1.73 & 68.31 $\pm$ 1.15 \\
ProtoNet~\cite{snell2017prototypical} & 59.02 $\pm$ 1.00 & 31.86 $\pm$ 1.02 & 61.56 $\pm$ 1.09 & 34.12 $\pm$ 1.06 & 66.47 $\pm$ 0.94 & 48.80 $\pm$ 1.12 & 79.49 $\pm$ 0.94 & 79.19 $\pm$ 0.78 \\
MatchingNet~\cite{vinyals2016matching} & 62.35 $\pm$ 1.06 & 28.50 $\pm$ 0.93 & \textbf{65.41 $\pm$ 1.06} & 30.16 $\pm$ 0.94 & 70.50 $\pm$ 0.99 & 44.01 $\pm$ 1.03 & 85.44 $\pm$ 0.63 & 84.10 $\pm$ 0.70 \\
fo-MAML~\cite{finn2017model} & 56.04 $\pm$ 1.08 & 20.64 $\pm$ 0.76 & 58.01 $\pm$ 1.12 & 21.24 $\pm$ 0.81 & 58.81 $\pm$ 1.12 & 32.38 $\pm$ 0.96 & 79.12 $\pm$ 0.95 & 71.88 $\pm$ 1.00 \\
fo-Proto-MAML~\cite{triantafillou2019meta} & 60.98 $\pm$ 1.01 & 28.32 $\pm$ 0.91 & 63.18 $\pm$ 1.02 & 29.08 $\pm$ 0.93 & 71.44 $\pm$ 1.00 & 46.98 $\pm$ 1.07 & 89.21 $\pm$ 0.70 & 86.70 $\pm$ 0.71 \\
CTX~\cite{doersch2020crosstransformers} & 56.29 $\pm$ 0.93 & 26.93 $\pm$ 0.91 & 58.52 $\pm$ 0.92 & 27.40 $\pm$ 0.95 & 63.00 $\pm$ 1.01 & 44.11 $\pm$ 1.06 & 94.51 $\pm$ 0.39 & \textbf{92.06 $\pm$ 0.48} \\
CTX+SimCLR~\cite{doersch2020crosstransformers} & \textbf{62.70 $\pm$ 1.07} & \textbf{38.56 $\pm$ 1.12} & 64.86 $\pm$ 1.09 & \textbf{38.22 $\pm$ 1.07} & \textbf{73.11 $\pm$ 0.93} & \textbf{61.47 $\pm$ 1.11} & \textbf{89.23 $\pm$ 0.60} & 90.01 $\pm$ 0.49 \\

\hline

\end{tabular}}
\caption{Benchmarking results on few-shot object classification in terms of 95\% confidence intervals for classification accuracy with \emph{episodic testing} consisting of 600 episodes as in Meta-Dataset~\cite{triantafillou2019meta}.}
\label{table:few-shot-classification}
\vspace{-7mm}
\end{table*}

\subsection{Training and Testing}

Our goal in building this dataset is to develop perception models that can classify objects in cluttered scenes with few examples per class. Therefore, we reserve the objects in the OCID dataset for testing, which are real-world objects in cluttered scenes (Fig.~\ref{fig:segmentation}). For training, we use the synthetic images rendered using Google Scanned Objects. The reasons for using synthetic data for training are: i) it is easy to generate cluttered scenes with ground truth annotations; ii) we can study sim-to-real transfer with our dataset.

In few-shot learning or meta-learning settings, training and testing have a support and query set. During training, the labels of the support and query sets are available. So we can use the labels of the query set to compute the training loss. During testing, labels of the support set are provided and the goal is to infer labels of the query set. Testing classes can be different from the training classes. When using our dataset, we consider two types of support sets: \emph{clean support sets} and \emph{cluttered support sets}. Clean support sets only consist of images of clean backgrounds without occlusions, while cluttered support sets contain images with different backgrounds and occlusions. Fig.~\ref{fig:support_query} illustrates these two types of support sets and their query sets. Since query images are from cluttered scenes, training with clean support sets is more challenging. However, if a method can work well with clean support sets, it requires less annotations, i.e., no annotation from cluttered scenes is needed. For example, given a novel object, we can just collect a few images of the object in a clean background in order to recognize it. Therefore, we encourage models that can perform well with clean support sets in our dataset.



\section{Benchmarking Experiments}
\label{sec:result}

In this section, we evaluate the state-of-the-art methods for few-shot learning and meta-learning on our dataset.
	
\subsection{Few-Shot Object Classification}

First, we follow the training and testing procedure designed in Meta-Dataset~\cite{triantafillou2019meta} and evaluated the following methods on our dataset: $k$-NN baseline that classifies each query example to the class of its closest support example, Finetune baseline that trains a classifier on top of the feature embedding using the support set of a test episode, Prototypical Networks~\cite{snell2017prototypical}, Matching Networks~\cite{vinyals2016matching}, first-order Model Agnostic Meta-Learning (fo-MAML)~\cite{finn2017model}, first-order Proto-MAML (fo-Proto-MAML) introduced in~\cite{triantafillou2019meta} and CrossTransformers (CTX)~\cite{doersch2020crosstransformers}. CTX uses an attention mechanism to compute a `query-aligned' prototype for each class, and then use these prototypes as in Prototypical Networks. \cite{doersch2020crosstransformers} also introduces using SimCLR~\cite{chen2020simple} as training episodes by treating every image as its own class for self-supervised learning (CTX+SimCLR). Details about the training and testing setup can be found in the supplementary materials. 95\% confidence intervals for the few-shot classification accuracy of these methods are presented in Table~\ref{table:few-shot-classification}. 

Model training is performed on 112 classes of our synthetic dataset, and testing is conducted on 52 classes in the OCID dataset~\cite{suchi2019easylabel} and 13 validation classes in the synthetic dataset. The backbones of these methods are ResNet-34 except for the Finetune baseline (ResNet-18 due to GPU memory limit). As in Meta-Dataset~\cite{triantafillou2019meta}, we compare with and without pre-training for the backbone network, where pre-training initializes the backbone weights as the $k$-NN Baseline model trained on ImageNet. We also use either clean support sets or cluttered support sets during training. The choice of pre-training and support set generates 4 training settings in Table~\ref{table:few-shot-classification}. For testing episodes, we evaluate on both clean support sets and cluttered support sets.

From the results in Table~\ref{table:few-shot-classification}, we have the following observations. i) Adding pre-training is beneficial. Most classification accuracies are improved with pre-training. ii) The performance on the synthetic classes is much better than on the real classes. We can see the sim-to-real gap clearly. iii) Using cluttered support sets can achieve better performance than using clean support sets because query sets contain different backgrounds and occlusions. However, obtaining annotations for cluttered support sets in the real world is expensive. Methods using clean support sets in testing are encouraged. iv) Among the 52 classes in OCID, we separately tested on 41 unseen classes, i.e., novel classes not presented in training and 11 seen classes. Overall, the performance on seen classes is better. v) Among these evaluated methods, CrossTransformers~\cite{doersch2020crosstransformers} achieves the best performance, which is consistent on other few-shot learning datasets~\cite{triantafillou2019meta}. CTX+SimCLR has a large margin when using clean support sets compared to other methods, highlighting the importance of self-supervised representation learning in SimCLR~\cite{chen2020simple}. This experiment suggests that pre-training and self-supervised contrastive representation learning can be critical in few-shot learning and meta-learning.

\begin{table*}[t]
\centering
\resizebox{0.75\linewidth}{!}{\begin{tabular}{|c|ccc|ccc|}
\hline
\multirow{4}{*}{Method} &   \multicolumn{6}{c|}{OCID (Real)~\cite{suchi2019easylabel}} \\ \cline{2-7}
& \multicolumn{3}{c|}{Use GT segmentation (\#classes, \#objects)} & \multicolumn{3}{c|}{Use segmentation from~\cite{xiang2021learning} (\#classes, \#objects)} \\ \cline{2-7}
& All (52, 2300) & Unseen (41, 1598) & Seen (11, 702) & All (52, 2300) & Unseen (41, 1598) & Seen (11, 702) \\ & Clean S & Clean S & Clean S &  Clean S & Clean S & Clean S \\ \hline \multicolumn{7}{|c|}{Training setting: clean support set with pre-training (top-1, top-5)} \\ \hline
$k$-NN~\cite{triantafillou2019meta}  & 14.65, 25.22  & 15.33, 24.41 & 41.03, 72.65 & 12.70, 23.22 & 13.70, 22.59  & 36.75, 67.95  \\
Finetune~\cite{triantafillou2019meta} & 22.26, 50.17 & \textbf{26.41}, 58.20 & 31.62, 80.34 & 21.30, 48.57 & \textbf{24.34}, 53.94 & 35.47, 67.38 \\
ProtoNet~\cite{snell2017prototypical} & \textbf{25.17}, \textbf{57.30} & 25.22, \textbf{58.45} & 51.99, \textbf{94.73} & \textbf{22.96}, \textbf{51.96} & 22.65, \textbf{54.32} & 49.86, \textbf{87.75}  \\
MatchingNet~\cite{vinyals2016matching} & 17.39, 48.35 & 14.64, 50.06 &  51.85, 90.31  & 15.78, 45.13 & 13.08, 46.93 & 49.15, 84.47 \\
fo-MAML~\cite{finn2017model} & 11.43, 31.48 & 11.58, 34.73 & 36.89, 69.94 & 10.91, 29.17 & 10.01, 32.35 & 31.77, 63.68 \\
fo-Proto-MAML~\cite{triantafillou2019meta} & 14.35, 28.96 & ~~5.63, 40.61 & 45.58, 71.51 & 13.39, 26.96 & ~~5.51, 37.73 & 41.74, 67.24\\
CTX~\cite{doersch2020crosstransformers} & 17.48, 46.57 &  18.21, 49.81 & 51.85, 87.75 & 15.70, 43.83 & 16.90, 46.31 & 47.86, 81.34   \\
CTX+SimCLR~\cite{doersch2020crosstransformers} & 18.57, 50.30 & 20.46, 51.06 & \textbf{57.55}, 93.16 & 16.48, 46.17 & 17.71, 47.12 & \textbf{52.14}, 85.75 \\ \hline
\multicolumn{7}{|c|}{Training setting: cluttered support set with pre-training (top-1, top-5)} \\ \hline
$k$-NN~\cite{triantafillou2019meta}  & 13.70, 23.83  & 15.33, 24.28 & 47.72, 72.79 & 13.26, 23.22  & 14.14, 22.90  & 44.73, 68.66  \\
Finetune~\cite{triantafillou2019meta} & 22.17, 53.35 & 24.34, 55.63 & 31.91, 71.51 & 18.26, 44.22 & 20.65, 52.00 & 36.04, 69.52 \\
ProtoNet~\cite{snell2017prototypical} & 21.35, 50.57 & 22.34, 51.31  & 51.99, 90.46 & 18.61, 47.22 & 18.21, 48.12 & 45.44, 85.33  \\
MatchingNet~\cite{vinyals2016matching} & 17.52, 50.96 & 17.77, 52.32 & 49.43, 88.18 & 16.52, 46.52 & 15.58, 48.81 & 43.45, 82.76 \\
fo-MAML~\cite{finn2017model} & 16.48, 38.52 & 13.70, 39.49 & 37.46, 77.07  & 15.35, 35.04 & 11.08, 34.36 & 40.31, 69.94 \\
fo-Proto-MAML~\cite{triantafillou2019meta} & 11.04, 28.70 & ~~4.01, 38.67 & 43.73, 72.65 & ~~9.91, 26.35 & ~~3.57, 35.79 & 40.46, 68.09 \\
CTX~\cite{doersch2020crosstransformers} & 19.00, 45.48 &  17.71, 44.74 & 51.85, 88.75 & 17.13, 42.22 & 16.08, 42.12 & 47.15, 83.19  \\
CTX+SimCLR~\cite{doersch2020crosstransformers} & \textbf{24.61}, \textbf{62.39} & \textbf{25.16}, \textbf{63.52} & \textbf{65.81}, \textbf{96.30} & \textbf{22.17}, \textbf{57.43} & \textbf{23.28}, \textbf{57.57} &  \textbf{59.12}, \textbf{88.32}\\ \hline


\multicolumn{7}{|c|}{Using pre-trained CLIP models~\cite{radford2021learning}} \\ \hline

Few-shot Tip-Adapter ViT-L/14-Finetune~\cite{tip_adapter_eccv22} & \textbf{60.17}, 83.04
 &	\textbf{59.64}, 85.17	 & 85.75, \textbf{99.00} & \textbf{54.87}, \textbf{78.91} &	\textbf{56.07}, 80.29 & 79.20, 91.88 \\

Few-shot Tip-Adapter ViT-L/14~\cite{tip_adapter_eccv22} & 56.78, 83.22 &	55.38, 84.86	 & \textbf{86.89}, 98.58 & 52.35, 76.26 & 51.69, 79.04	 & \textbf{80.06}, \textbf{92.45} \\

Zero-shot CLIP ViT-L/14~\cite{radford2021learning} &	54.57, \textbf{84.74}	 & 55.94, \textbf{87.92} & 83.62, 98.58 &		50.43, 78.52 & 52.07, \textbf{81.54} & 75.07, 92.17 \\
Zero-shot CLIP ViT-B/32~\cite{radford2021learning} &	41.87, 75.26 & 41.30, 77.91 & 78.06, 97.58 &		39.83, 69.43 & 39.17, 72.09 & 70.66, 90.88 \\
Zero-shot CLIP ViT-B/16~\cite{radford2021learning} &	40.70, 73.96 & 40.24, 76.03 & 76.50, 95.73  &		39.35, 68.83 & 38.61, 70.15 & 70.66, 88.89 \\
Zero-shot CLIP RN50x64~\cite{radford2021learning}	& 42.96, 75.83	 & 43.62, 77.41 & 76.64, 96.01 &		40.04, 70.87 & 41.74, 72.22 & 69.94, 90.46 \\
Zero-shot CLIP RN50x16~\cite{radford2021learning}	& 38.52, 73.04	 & 40.11, 75.72 & 79.49, 96.30 &		35.65, 67.30 & 37.30, 69.77 & 70.94, 89.74 \\
Zero-shot CLIP RN50x4~\cite{radford2021learning}	& 35.96, 68.52	 & 34.42, 70.03 & 73.93, 95.73 &		34.00, 63.78 & 32.48, 65.46 & 67.95, 88.60 \\
Zero-shot CLIP ResNet-101~\cite{radford2021learning}	& 32.96, 68.30	 & 32.67, 69.52 & 77.49, 96.87 &		31.09, 63.87 & 31.85, 65.96 & 69.66, 89.74 \\
Zero-shot CLIP ResNet-50~\cite{radford2021learning}	& 25.91, 58.43	 & 29.04, 64.39 & 61.40, 93.16 &		24.70, 55.61 & 28.04, 61.20 & 57.69, 86.47 \\
\hline

\end{tabular}}
\caption{Benchmarking results on joint object segmentation and few-shot classification in terms of top-1 and top-5 classification accuracy with \emph{non-episodic testing} on the OCID dataset~\cite{suchi2019easylabel}. For CLIP-based models, different image encoder backbones are tested: ResNet~\cite{he2016deep}, EfficientNet~\cite{tan2019efficientnet} style ResNet (RN50x4, RN50x16, RN50x64) and Vision Transformers (ViT-B/16, ViT-B/32, ViT-L/14)~\cite{dosovitskiy2020image}. }
\label{table:segmentation}
\vspace{-4mm}
\end{table*}

\subsection{Joint Object Segmentation and Few-Shot Classification}
In this experiment, we conduct non-episodic testing on all the 2,300 objects among the 52 classes in the OCID dataset. When cropping objects from the original images for classification, we tested using ground truth masks versus using the predicted masks from~\cite{xiang2021learning}. We need to assign a mask to each object when using predicted masks. This is achieved by the Hungarian method with pairwise F-measure that computes matching between predicted masks and ground truth objects. These cropped objects construct the query set. We use the real-world objects that we captured on a tabletop as the clean support sets. We present top-1 and top-5 classification accuracies of the evaluated methods in Table~\ref{table:segmentation}. Few-shot learning methods are trained on the 112 classes of our synthetic dataset with pre-training. In addition, we also tested the CLIP models~\cite{radford2021learning,tip_adapter_eccv22} with different image encoder backbones. If an object cannot be segmented by a segmentation method, i.e., no assigned mask for the Hungarian matching, we consider this object as a misclassification. In this way, the classification accuracy also accounts for the segmentation performance and using ground truth segmentation masks focuses on evaluating the classification performance only. From Table~\ref{table:segmentation}, we can see that: i) The top-1 accuracy is around 25\% for the best few-shot learning method trained on synthetic data, which indicates that there is still a large margin to be improved in this setting. The difficulties lie in using synthetic images for training and clean support sets during testing.  ii) Classification accuracies of seen classes are much higher than unseen classes. iii) Overall, CTX+SimCLR performs better when training with cluttered support sets, while Prototypical Network performs better when training with clean support sets. iv) The pre-trained CLIP models~\cite{radford2021learning} perform much better than trained few-shot learners on classifying these objects. We think it is due to the large scale real-world data that CLIP models are trained on. The Tip-Adapter~\cite{tip_adapter_eccv22} adapts CLIP models for few-shot classification. Its training-free model and the fine-tuned variant improve over the original CLIP models.

\begin{figure}
	\centering
	\includegraphics[width=.42\textwidth]{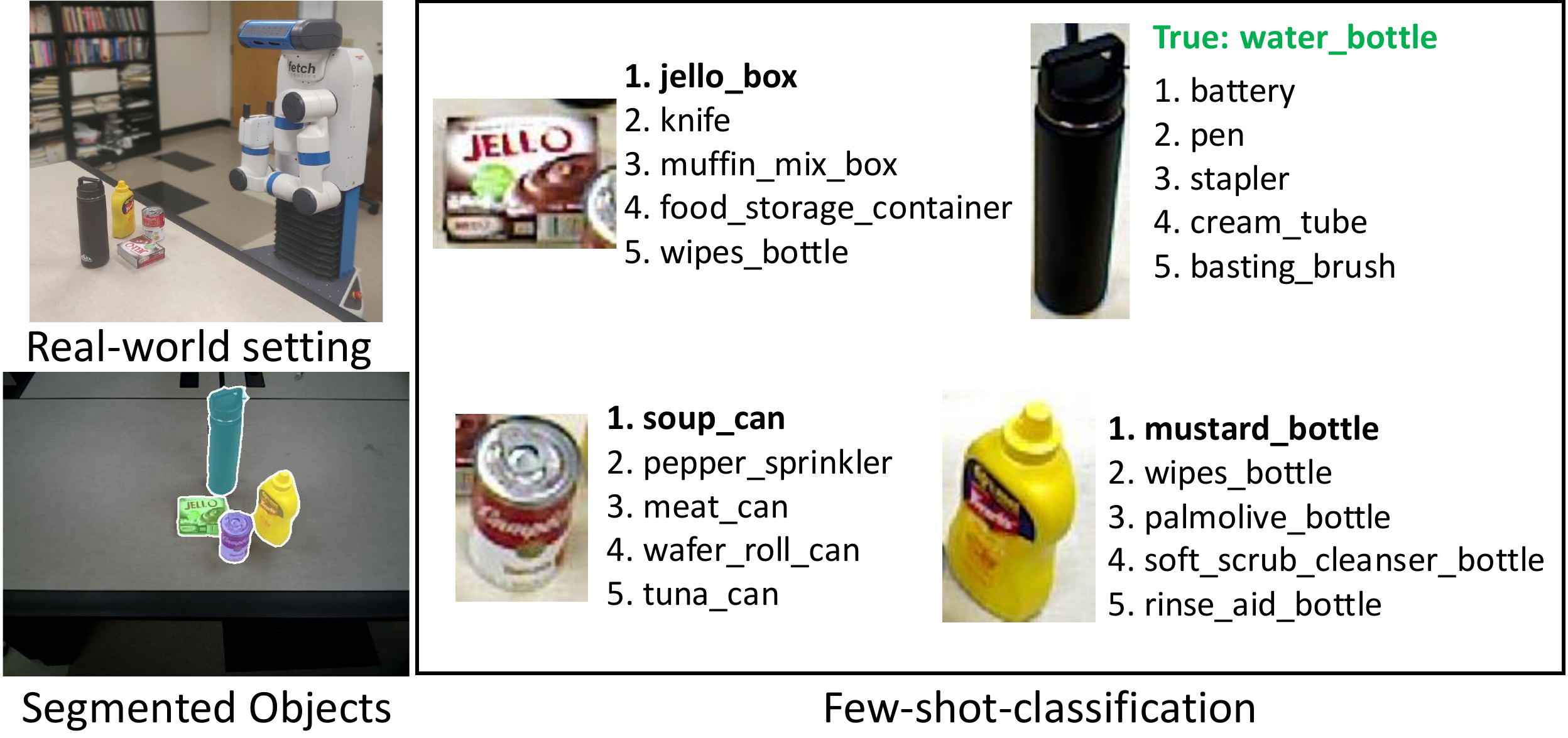}
	\caption{Qualitative results with top-5 predictions from our real-world testing with the fine-tuned Tip-Adapter (ViT-L/14) model~\cite{tip_adapter_eccv22}.}
	\label{fig:real_world}
	\vspace{-6mm}
\end{figure}

\subsection{Qualitative Results in the Real World}

In this experiment, we aim to build a few-shot classification model that works best on real-world perception systems. So we train CTX+SimCLR~\cite{doersch2020crosstransformers} with all the real and synthetic data in our dataset and then test the trained model in our lab. Cluttered support sets are used for training. RGB-D images are collected from a Fetch mobile manipulator and we used \cite{xiang2021learning} for object segmentation. We tested on 32 objects with 4 objects in an image. The CTX+SimCLR model achieves 28.13\% and 56.25\%, the pre-trained CLIP-ViT-L/14 model achieves 65.63\% and 81.25\% and the fine-tuned Tip-Adapter (ViT-L/14) model achieves 65.63\% and 84.38\% top-1 and top-5 accuracy respectively. Fig.~\ref{fig:real_world} shows one testing image and the classification results from the Tip-Adapter~\cite{tip_adapter_eccv22} model. Please see the supplementary video for these classification results. The low top-1 accuracy indicates the difficulty of the few-shot object classification problem in the real world. Most failure cases in few-shot classification are due to the differences between testing and training objects. For example, the black bottle in Fig.~\ref{fig:real_world} was not seen during training. How to achieve better generalization in few-shot object classification is an interesting direction to explore. We hope that our dataset can be used to build better models for this problem. 


\section{Conclusion and Future Work}
\label{sec:conclusion}

We introduce the Few-Shot Object Learning (\textsc{FewSOL}) dataset for few-shot object recognition. Different from existing datasets for few-shot learning, our dataset contains household objects such as personal items, tools and fruits. We provide RGB-D images, object segmentation masks, poses and attribute annotations in the dataset. We hope the dataset can facilitate progress on robot object perception. If a robot can recognize all the object classes in the dataset (198 classes of real objects), this will help lots of robotic applications such as manipulation, object retrieval, object grounding, task planning, and so on. We demonstrated using our dataset for (i) few-shot object classification and (ii) joint object segmentation and few-shot classification in this paper. The experimental results show a need to improve the few-shot recognition performance for real-world robotic applications. In the future, we plan to study how to leverage depth data and multi-view information to improve few-shot object classification. We also plan to study few-shot object representation learning for shape reconstruction, object pose estimation and object attribute recognition using the \textsc{FewSOL} dataset.



\section*{Acknowledgments}
This work was supported in part by the DARPA Perceptually-enabled Task Guidance (PTG) Program under contract number HR00112220005.



{
\bibliographystyle{IEEEtran}
\bibliography{example}  
}

\end{document}


\maketitle

\tableofcontents
\newpage

\section{Definitions} 
\label{sec: definitions}

A \textbf{clean support set} (Clean S) consists of single objects with clean background. 
A \textbf{cluttered support set} (Cluttered S) consists of single objects or objects from a collection of objects in a cluttered scene with different backgrounds and occlusions. 
An \textbf{episodic testing} indicates using a number of episodes in testing where each episode consists of several support and query sets from different classes.
A \textbf{non-episodic testing} indicates the classification of all the test images in the test set without using episodes. These images can be considered to be a whole query set.

\begin{figure}[H]
  \centering
  \includegraphics[width=\textwidth]{supp-figures/support_query.pdf}
  \caption{\{a,c\} and \{b,d\} are examples of clean and cluttered support sets of training and testing episodes, respectively.}
  \label{fig:clean-clutter-support}
\end{figure}

\section{Data and Evaluation Metrics} \label{sec: datasets-and-eval-metrics}
\subsection{Data Preparation Pipeline} \label{sec: data-pipeline}

We follow the same data preparation pipeline devised in META-DATASET~\citep{triantafillou2019meta}. Raw FewSOL dataset is converted to the TFRecord\footnote{\url{https://www.tensorflow.org/tutorials/load_data/tfrecord}} format. The data pipeline utilizes TFRecords to form episodes. The data pipeline randomly shuffles and selects entries from TFRecords to form the train and test episodes which are then used to train and test the respective models mentioned in Section~\ref{sec: models}.

\subsection{Dataset Split}

\paragraph{Train and Validation Sets.} For few-shot classification and joint object segmentation and few-shot classification, training data comprised 125 synthetic object classes. The training data is split into train and validation sets based on a 90:10 ratio of classes. The classes were first sorted in descending order based on the number of support examples. The top $90\%$ classes, i.e., 112 classes form the train set, and the last $10\%$, i.e., 13 classes form the validation set. This would allow using more variety of support examples for training and validation using fewer support examples which is an apt case for the real world as well. With this setup, it is clear that train and validation classes are disjoint. 

\paragraph{Test Sets.}\label{par: FewSOL dataset:variants}
For few shot classification, as shown in Table~\ref{tbl: test-data-variants}, four variants of the test data were generated: (i) \textbf{All} comprises of 52 test classes, (ii) \textbf{Unseen} comprises of 41 test classes (disjoint from training classes), (iii) \textbf{Seen} comprises of 11 test classes (common with training classes) and (iv) \textbf{Unseen (Synthetic)}  comprises of 13 test classes (common with the validation set portion of the training data). For joint object segmentation and few-shot classification, only \{All, Unseen, Seen\} variants were used.

\begin{wraptable}{t}{0.5\textwidth}
    \centering
    \begin{tabular}{cc}
    \hline
    \textbf{Variant} & \textbf{\#Classes}  \\ \hline
    All              & 52                     \\
    Unseen           & 41                                 \\
    Seen             & 11                                \\
    Unseen (Synthetic) & 13       \\ \hline                     
    \end{tabular}
    \caption{Test data variants.}
    \label{tbl: test-data-variants}
\end{wraptable}

The Seen variant has 11 real object classes that coincide with synthetic object classes, hence a need to keep all 11 synthetic object classes in the train split came up. This is useful for studying the effect on these classes during episodic testing of the Seen variant. This need was catered by transferring any of the 11 synthetic object classes common with the Seen variant test classes from validation split to train split. The void of the transferred classes in the validation split was filled by the bottom classes of the initially sorted train split so that classes with fewer support samples stay in the validation split and the train split gets classes with more support samples which may aid in training the models. Fortunately, with the current setup of $112/13$ after sorting only one class had to be swapped. The last train split class \textbf{product\_box} was swapped with \textbf{sponge} class from validation split. After all these steps, validation classes were product\_box, eagle, wood\_tower, cream\_box, cake\_pan, screwdriver, food\_can, bottle, paper\_roll, honey\_dipper, racoon, hard\_drive, and rubber\_band. Furthermore, the last step of image filtration discussed in Section~\ref{par: discard-img-15px-rule} was performed, i.e., all images with either width or height less than 15 pixels were discarded as it may cause the problem shown in Figure~\ref{fig:img-filter-inspiration}.

For the real-world setup, all real and synthetic data from the FewSOL dataset were used for training. Hence, the training data comprised 323 (125+198) classes. This forms a mixture of synthetic and real object classes. Since only 52 out of 198 classes had query sets, the query sets for the remaining 146 classes were generated by splitting the available support set images into two halves. One half was considered as the support set while the other formed the query set. A 90:10 split yielded 291 train and 32 validation classes. While these 32 validation classes have real images, there are similar 32 synthetic classes in 291 train classes. This is a good setup as the Sim-to-Real transfer effect could be utilized here. Thus, in this way train and validation classes are disjoint. 

\subsection{Data preprocessing}
\label{sec:data-preprocessing}
\paragraph{Discard Low Resolution Images.}
\label{par: discard-img-15px-rule}

\begin{wrapfigure}{r}{0.5\textwidth}
  \centering 
  \includegraphics[scale=3.5]{supp-figures/too-short-problem.pdf}
  \caption{The left (low resolution and blurred) image shows an object after cropping from a cluttered scene while the right image shows the same when resized to 84x84x3 size. This may not help during training as all the images are reshaped to 84x84x3 or 126x126x3 before being fed into any model for few-shot learning.}
  \label{fig:img-filter-inspiration}
    \vspace{-7mm}
\end{wrapfigure}

As shown in Figure~\ref{fig:img-filter-inspiration}, low-resolution images, i.e., images having very small width or height when resized to 84x84x3 or 126x126x3 yields blurred images with no useful feature to learn. Hence, we discard images with width or height less than 15 pixels to avoid including noises while training and testing (except during real-world testing with 198 classes and joint object segmentation and few-shot classification training and testing).  The threshold for 15 pixels has been chosen with the heuristic that it is very small and going less than that may not help. Also, increasing this threshold might discard a lot of images that could be useful for training and testing.


\paragraph{Support Set Oversampling.}
\label{par: support-set-oversampling} 

This step is required for few shot classification only. As discussed in Section~\ref{sec: data-pipeline}, the data pipeline randomly shuffles and selects entries from TFRecords to form episodes. This affects FewSOL dataset as it has fewer support samples w.r.t. query (refer Table~\ref{tbl:synthetic-dataset} and \ref{tbl:real+ocid-dataset}). Hence, while performing the filtration discussed in Section~\ref{sec: changes-to-md-codebase}, either no or fewer support images might be left. This is a case of set imbalance. Thus with the standard data pipeline of META-DATASET~\citep{triantafillou2019meta}, the solution was to oversample the support images to match the cardinality of the query set. This oversampling only happens at the TFRecords level and not on the raw images. This gives a fair chance to the support samples to be selected alongside the query samples. Oversampling is done as 
$N_{ic} = \ceil{Q_{c}/S_{c}}$, where $N_{ic}$ is the number of times to oversample the $i^{th}$ support image in class \textit{C}, $S_{c}$ and $Q_{c}$ are the number of available support and query images in class \textit{C}, respectively.



\subsection{Evaluation Metrics}
For few shot classification similar to META-DATASET~\citep{triantafillou2019meta}, we report 95\% confidence intervals for classification accuracy with episodic testing consisting of 600 episodes. This is reasonable as variable ``N-way K-shot'' setup is used to construct test episodes. For joint object segmentation and few-shot classification along with real-world testing, top-1 and top-5 classification accuracy are reported on all query data while utilizing all support data.




\section{Experiment Details}
\label{sec:training-and-testing-details}

\subsection{Models}
\label{sec: models}
We purposefully restricted our experiments to a selected set of few-shot and meta-learning methods from the growing list of META-DATASET~\citep{triantafillou2019meta}, viz., $k$-NN~\cite{triantafillou2019meta}, $k$-NN-Finetuned~\cite{triantafillou2019meta}, Prototypical Networks~\cite{snell2017prototypical}, Matching Networks~\cite{vinyals2016matching}, Model Agnostic Meta-Learning (MAML)~\cite{finn2017model}, Proto-MAML~\cite{triantafillou2019meta}, CrossTransformers (CTX)~\cite{doersch2020crosstransformers} and CTX+SimCLR~\cite{doersch2020crosstransformers} variant due to the combination explosion of the experiments, hardware constraints and stability of META-DATASET~\citep{triantafillou2019meta} codebase. We think that these methods are a good representative sample to begin with.

\subsection{Embedding Networks (Backbones)}
\label{sec: backbones}
Each model described in Section~\ref{sec: models} needs a backbone to learn useful features. META-DATASET~\citep{triantafillou2019meta} offers a variety of backbone options: four layer convolutional network (four\_layer\_convnet), resnet18 (resnet), wide\_resnet, and resnet34. All models use resnet34 backbone architecture except $k$-NN-Finetuned~\cite{triantafillou2019meta} which uses resnet due to GPU memory limit. Resnet34 was mainly chosen due to its superior performance during pretraining using ImageNet~\citep{deng2009imagenet}. Resnet with image\_size=84 gave on par performance w.r.t. Resnet34, thus it was the next substitute for $k$-NN-Finetuned~\cite{triantafillou2019meta}. CTX uses resnet34 with an extra setting of max\_stride=16.

\begin{table*}[t]
\centering
\begin{tabular}{ccc}
\hline
\textbf{Backbone}    & \textbf{Accuracy}           & \textbf{Image Size} \\ \hline
four\_layer\_convnet & 25.30\% & 84                  \\
four\_layer\_convnet & 31.17\% & 126             \\\hline
resnet               & 33.20\% & 84                  \\
resnet               & 29.96\%                        & 126   \\\hline
resnet34             & \textbf{34.44\%}                        & 84                  \\
resnet34             & \textbf{33.76\%}                        & 126                \\ \hline
resnet34 (max-stride=16)             & \textbf{35.14\%}                        & 126                \\
\hline
\end{tabular}
\caption{Ablation study of image size w.r.t. few-shot classification accuracy for selected backbones using $k$-NN~\citep{triantafillou2019meta} as the learner model. Here, $k$-NN~\citep{triantafillou2019meta} is trained on ImageNet~\citep{deng2009imagenet}. $k$-NN~\citep{triantafillou2019meta} was selected following the suggestion from META-DATASET~\citep{triantafillou2019meta} that some of the best meta-learning models are initialized from the weights of a batch baseline.}
\label{tbl: ablation-study-backbone-pretraining}
\end{table*}

\subsection{Backbone Pretraining}
\label{sec: backbone-pretraining}
As in META-DATASET~\citep{triantafillou2019meta}, we pretrain the backbones mentioned in Table~\ref{tbl: ablation-study-backbone-pretraining} using ImageNet~\citep{deng2009imagenet} for 50K updates. These backbones were selected based on the suggestions from META-DATASET~\citep{triantafillou2019meta}. An ablation study on the effect of image size on few shot classification accuracy of selected backbones is shown in Table~\ref{tbl: ablation-study-backbone-pretraining}. The accuracy reported is the mean component of the 95\% confidence interval. Confidence interval is used mainly due to the variable ``N-way K-shot'' setup. It is clear from Table~\ref{tbl: ablation-study-backbone-pretraining} that resnet34 performs better. Thus it was selected as the backbone. Since CTX+SimCLR~\citep{doersch2020crosstransformers} performs better at higher resolution, we chose to go ahead with the image\_size=126. We tried using image\_size=224 but were constrained due to the GPU memory limit. To maintain uniformity and perform comparison across various methods, resnet34 was used as the backbone with image\_size=126. The GPU memory limit in case of $k$-NN-Finetuned~\cite{triantafillou2019meta} made the option of going ahead with resnet as the backbone with image\_size=84 as it performed on par with resnet34 (refer Table~\ref{tbl: ablation-study-backbone-pretraining}). The images are resized to (image\_size x image\_size) before the forward pass similar to META-DATASET~\citep{triantafillou2019meta}.

\begin{table}[]
\centering
\begin{tabular}{ccccc}
\hline
\textbf{Model} & \textbf{Backbone}  &    \textbf{Max Stride}     & \textbf{Image Size}     & \textbf{Num Training Updates}\\ \hline
  $k$-NN~\cite{triantafillou2019meta}       &  resnet34 &    -   & 126 & 75K   \\
  $k$-NN-Finetuned~\cite{triantafillou2019meta}     & resnet          &   -  & 84 & 75K    \\
  ProtoNet~\cite{snell2017prototypical}     &  resnet34 &    -   & 126 & 75K     \\
  MatchingNet~\cite{vinyals2016matching}    & resnet34 &    -   & 126    & 75K  \\
  fo-MAML~\cite{finn2017model}      & resnet34 &    -   & 126  & 75K    \\
  fo-Proto-MAML~\cite{triantafillou2019meta}    & resnet34 &    -   & 126    & 75K  \\
  CTX~\cite{doersch2020crosstransformers}   & resnet34 &    16   & 126  & 100K    \\
  CTX+SimCLR~\cite{doersch2020crosstransformers} & resnet34 &    16   & 126   & 400K    \\ \hline
\end{tabular}
\caption{Details of hyperparameters for training and testing. Num Training Updates denotes the maximum number of updates that training process can perform.}
\label{tbl: train-test-backbone-imsize-updates}
\end{table}

\subsection{Hyperparameters}
The best hyperparameters reported for each model in META-DATASET~\citep{triantafillou2019meta} have been used for our experiments. The changes to the default hyperparameters related to backbones are discussed in Section~\ref{sec: backbone-pretraining}. Following are the changes from our end: (i) Model training involving resnet34 backbone have learning\_rate set as 0.001052178216688174, (ii) Number of validation episodes are set to 60 (default 600) taking into consideration the minimum need for constructing a confidence interval and the exceeding time for validation, (iii) EpisodeDescriptionConfig.max\_support\_size\_contrib\_per\_class is set to 9 instead of default 500, (iv) EpisodeDescriptionConfig.max\_ways\_upper\_bound is set to 52 and 198 for few shot classification and joint object segmentation and few-shot classification respectively.

\subsection{Training Details} 
\label{sec: training}

We have two training setups: (a) using 125 synthetic classes and (b) using all classes of the FewSOL dataset, i.e. 323 classes. The training setup (a) is further divided into two more variants based on the type of support set used for training: (i) using clean support sets and (ii) using cluttered support sets. Figure~\ref{fig:clean-clutter-support} shows the clean and cluttered support sets for few shot classification training and testing. Each support set setup is further divided into 2 types based on (i) using pretrained backbones and (ii) using backbones initialized randomly. Details of the backbones are mentioned in Section~\ref{sec: backbones}.  We use the same protocol as in META-DATASET~\citep{triantafillou2019meta} for episode composition. We use the variable ``N-way K-shot'' setup for training. Training setup (a) is used to train models mentioned in Section~\ref{sec: models}.

Training setup (b) uses all synthetic and real data from the FewSOL dataset to train 
CTX+SimCLR~\cite{doersch2020crosstransformers}. Only cluttered support sets are used for training this particular model which would be used for real-world testing. Refer Section~\ref{sec:4.3-qual-results-in-real-world} for more details.

\subsection{Selection of the Best Trained Checkpoints}
\label{sec: select-best-ckpt}
As stated in META-DATASET~\citep{triantafillou2019meta}, actual early stopping is not performed, in that training is not stopped early according to validation performance. Instead, checkpoints are recorded every $500^{th}$ update during training, and validation error is saved at these times. The checkpoint corresponding to the least validation error is chosen as the best one. This procedure is used for selecting the best-trained model checkpoints in all the experiments.

\subsection{Testing Details} 
\label{sec: testing}
For the few shot classification experiment, we use the variable ``N-way K-shot'' setup as in META-DATASET~\citep{triantafillou2019meta} for constructing test episodes. We call this `episodic testing'. Testing was done on all the variants using the models trained on 4 training setups discussed in Section~\ref{sec: training}. For joint object segmentation and few-shot classification experiment, we used all the query images from \{All, Unseen, Seen\} variants. Refer Section~\ref{par: FewSOL dataset:variants}. We call this the `non-episodic' setup. Few shot classification and joint object segmentation and few-shot classification testing share the same trained model checkpoints. They only differ in the type of testing. The former uses episodic whereas the latter uses non-episodic testing.

For real-world testing, we have setup an environment where a Fetch mobile manipulator is facing a desk containing various real-world objects mentioned in Section~\ref{fig:8-object-sets}. We have 8 setups each containing 4 different objects. Fetch takes an image containing the objects on the desk then \citep{xiang2021learning} is used for object segmentation. The object masks are then used to crop the detected objects. The cropped object images become the query set whereas the support set of 198 real classes forms the support set of the new real-world test set. This new real-world test set is tested using the model trained using training setup (b) (refer Section~\ref{sec: training}).

\subsection{Experimental Hardware} \label{sec:experimental-setup}
All experiments were conducted on two NVIDIA RTX A5000 24GB GPUs. A Fetch mobile manipulator\footnote{\url{https://fetchrobotics.com/fetch-mobile-manipulator}} was used for real-world testing.

\subsection{Changes introduced in META-DATASET~\citep{triantafillou2019meta} code}
\label{sec: changes-to-md-codebase}

\begin{wrapfigure}{r}{0.4\textwidth}
	\centering
	\includegraphics[width=0.4\textwidth]{supp-figures/4.3/fetch-facing-objects.jpg}
	\caption{Fetch mobile manipulator facing the objects from Set-1 (Figure~\ref{fig:4.3-set-1}) on a table}
	\label{fig:fetch-facing-objects}
\end{wrapfigure}

All the images are first converted into tfrecords format. The default tfrecords are a collection of example protocol buffer\footnote{\url{https://developers.google.com/protocol-buffers}} strings. Each example string consists of two elements, (i) image and (ii) label in bytes format. Meta-Dataset leverages data from 10 different datasets: ILSVRC-2012 (ImageNet~\cite{russakovsky2015imagenet}), Omniglot~\cite{lake2015human}, Aircraft~\cite{maji2013fine}, CUB-200-2011 (Birds~\cite{wah2011caltech}), Describable Textures~\cite{cimpoi2014describing},
Quick Draw~\cite{Jongejan2016quick}, Fungi~\cite{sulc2020fungi}, VGG Flower~\cite{nilsback2008automated}, Traffic Signs~\cite{houben2013detection} and MSCOCO~\cite{lin2014microsoft}. As we can see, unlike the FewSOL dataset, these data are not divided into support and query sets specifically. Hence, we added a new key to each example string namely \textbf{set} which could take one value from \{\textit{support, query, `'}\}. `' denotes that the particular example can belong to either support or query. This ensures the existing 10 datasets of META-DATASET~\citep{triantafillou2019meta} are compatible with the new protocol buffer structure. The default META-DATASET~\citep{triantafillou2019meta} pipeline shuffles and randomly selects a set of example strings which are then again randomly divided into support and query sets. This is reasonable as there is no bifurcation of support and query sets in the existing META-DATASET~\citep{triantafillou2019meta} datasets. But FewSOL dataset has a clear distinction between support and query and thus a need to select support and query data from the respective category was necessary. For this purpose, a perform\_filtration feature is introduced. If set to boolean `True', the pipeline filters the support and query sets so that the resulting data belongs to the respective categories. If set to boolean `False', the standard META-DATASET~\citep{triantafillou2019meta} data pipeline is run.

\clearpage
\section{All results}\label{sec: additional-results}

\begin{table}[H]
\centering
\resizebox{\linewidth}{!}{\begin{tabular}{|c|ccc|ccc|}
\hline
\multirow{4}{*}{Method} &   \multicolumn{6}{c|}{OCID (Real)~\cite{suchi2019easylabel}} \\ \cline{2-7}
& \multicolumn{3}{c|}{Use GT segmentation (\#classes, \#objects)} & \multicolumn{3}{c|}{Use segmentation from~\cite{xiang2021learning} (\#classes, \#objects)} \\ \cline{2-7}
& All (52, 2300) & Unseen (41, 1598) & Seen (11, 702) & All (52, 2300) & Unseen (41, 1598) & Seen (11, 702) \\ & Clean S & Clean S & Clean S &  Clean S & Clean S & Clean S \\ \hline \multicolumn{7}{|c|}{Training setting: clean support set with pre-training (top-1, top-5)} \\ \hline
$k$-NN~\cite{triantafillou2019meta}  & 14.65, 25.22  & 15.33, 24.41 & 41.03, 72.65 & 12.70, 23.22 & 13.70, 22.59  & 36.75, 67.95  \\
Finetune~\cite{triantafillou2019meta} & 22.26, 50.17 & \textbf{26.41}, 58.20 & 31.62, 80.34 & 21.30, 48.57 & \textbf{24.34}, 53.94 & 35.47, 67.38 \\
ProtoNet~\cite{snell2017prototypical} & \textbf{25.17}, \textbf{57.30} & 25.22, \textbf{58.45} & 51.99, \textbf{94.73} & \textbf{22.96}, \textbf{51.96} & 22.65, \textbf{54.32} & 49.86, \textbf{87.75}  \\
MatchingNet~\cite{vinyals2016matching} & 17.39, 48.35 & 14.64, 50.06 &  51.85, 90.31  & 15.78, 45.13 & 13.08, 46.93 & 49.15, 84.47 \\
fo-MAML~\cite{finn2017model} & 11.43, 31.48 & 11.58, 34.73 & 36.89, 69.94 & 10.91, 29.17 & 10.01, 32.35 & 31.77, 63.68 \\
fo-Proto-MAML~\cite{triantafillou2019meta} & 14.35, 28.96 & ~~5.63, 40.61 & 45.58, 71.51 & 13.39, 26.96 & ~~5.51, 37.73 & 41.74, 67.24\\
CTX~\cite{doersch2020crosstransformers} & 17.48, 46.57 &  18.21, 49.81 & 51.85, 87.75 & 15.70, 43.83 & 16.90, 46.31 & 47.86, 81.34   \\
CTX+SimCLR~\cite{doersch2020crosstransformers} & 18.57, 50.30 & 20.46, 51.06 & \textbf{57.55}, 93.16 & 16.48, 46.17 & 17.71, 47.12 & \textbf{52.14}, 85.75 \\ \hline
\multicolumn{7}{|c|}{Training setting: cluttered support set with pre-training (top-1, top-5)} \\ \hline
$k$-NN~\cite{triantafillou2019meta}  & 13.70, 23.83  & 15.33, 24.28 & 47.72, 72.79 & 13.26, 23.22  & 14.14, 22.90  & 44.73, 68.66  \\
Finetune~\cite{triantafillou2019meta} & 22.17, 53.35 & 24.34, 55.63 & 31.91, 71.51 & 18.26, 44.22 & 20.65, 52.00 & 36.04, 69.52 \\
ProtoNet~\cite{snell2017prototypical} & 21.35, 50.57 & 22.34, 51.31  & 51.99, 90.46 & 18.61, 47.22 & 18.21, 48.12 & 45.44, 85.33  \\
MatchingNet~\cite{vinyals2016matching} & 17.52, 50.96 & 17.77, 52.32 & 49.43, 88.18 & 16.52, 46.52 & 15.58, 48.81 & 43.45, 82.76 \\
fo-MAML~\cite{finn2017model} & 16.48, 38.52 & 13.70, 39.49 & 37.46, 77.07  & 15.35, 35.04 & 11.08, 34.36 & 40.31, 69.94 \\
fo-Proto-MAML~\cite{triantafillou2019meta} & 11.04, 28.70 & ~~4.01, 38.67 & 43.73, 72.65 & ~~9.91, 26.35 & ~~3.57, 35.79 & 40.46, 68.09 \\
CTX~\cite{doersch2020crosstransformers} & 19.00, 45.48 &  17.71, 44.74 & 51.85, 88.75 & 17.13, 42.22 & 16.08, 42.12 & 47.15, 83.19  \\
CTX+SimCLR~\cite{doersch2020crosstransformers} & \textbf{24.61}, \textbf{62.39} & \textbf{25.16}, \textbf{63.52} & \textbf{65.81}, \textbf{96.30} & \textbf{22.17}, \textbf{57.43} & \textbf{23.28}, \textbf{57.57} &  \textbf{59.12}, \textbf{88.32}\\ \hline

\multicolumn{7}{|c|}{\textcolor{blue}{\textbf{Zero-shot}} using pre-trained CLIP~\cite{radford2021learning} models (top-1, top-5)} \\ \hline
ViT-L/14@336px &	53.39, 83.30	 & 54.13, 85.73 &  82.62, 98.43 &	49.17, 77.22 & 50.06, 80.35 & \textbf{75.50}, \textbf{92.31} \\
ViT-L/14 &	\textbf{54.57}, \textbf{84.74}	 & \textbf{55.94}, \textbf{87.92} & \textbf{83.62}, \textbf{98.58} &		\textbf{50.43}, \textbf{78.52} & \textbf{52.07}, \textbf{81.54} & 75.07, 92.17 \\
ViT-B/32 &	41.87, 75.26 & 41.30, 77.91 & 78.06, 97.58 &		39.83, 69.43 & 39.17, 72.09 & 70.66, 90.88 \\
ViT-B/16 &	40.70, 73.96 & 40.24, 76.03 & 76.50, 95.73  &		39.35, 68.83 & 38.61, 70.15 & 70.66, 88.89 \\
RN50x64	& 42.96, 75.83	 & 43.62, 77.41 & 76.64, 96.01 &		40.04, 70.87 & 41.74, 72.22 & 69.94, 90.46 \\
RN50x16	& 38.52, 73.04	 & 40.11, 75.72 & 79.49, 96.30 &		35.65, 67.30 & 37.30, 69.77 & 70.94, 89.74 \\
RN50x4	& 35.96, 68.52	 & 34.42, 70.03 & 73.93, 95.73 &		34.00, 63.78 & 32.48, 65.46 & 67.95, 88.60 \\
ResNet-101	& 32.96, 68.30	 & 32.67, 69.52 & 77.49, 96.87 &		31.09, 63.87 & 31.85, 65.96 & 69.66, 89.74 \\
ResNet-50	& 25.91, 58.43	 & 29.04, 64.39 & 61.40, 93.16 &		24.70, 55.61 & 28.04, 61.20 & 57.69, 86.47 \\
\hline
\multicolumn{7}{|c|}{\textbf{Few-shot} using Tip-Adapter~\cite{tip_adapter_eccv22} models with hyperparameter search enabled, image\_size: 224x224, \textcolor{cyan}{\textit{\textbf{F}}}:   finetuned (top-1, top-5)} \\ \hline
ViT-L/14 & 56.78, \textbf{83.22} &	55.38, 84.86	 & \textbf{86.89}, 98.58 & 52.35, 76.26 & 51.69, 79.04	 & \textbf{80.06}, \textbf{92.45} \\
ViT-L/14-\textcolor{cyan}{\textit{\textbf{F}}} & \textbf{60.17}, 83.04 &	\textbf{59.64, 85.17}	 & 85.75, \textbf{99.00} & \textbf{54.87}, \textbf{78.91} &	\textbf{56.07}, \textbf{80.29} & 79.20, 91.88 \\
ViT-B/32 &	41.48, 77.48	 & 42.12, 78.60  & 81.20, 98.72  & 39.52, 71.52	& 39.36, 72.97 &  74.22, 91.74\\
ViT-B/32-\textcolor{cyan}{\textit{\textbf{F}}} & 45.48, 76.65	&	48.44, 79.47 &  81.77, \textbf{99.00} & 43.13, 68.30 &	43.37, 72.90 &  73.50, 91.03\\
ViT-B/16 &	47.00, 78.43	 & 44.93, 78.97 & 84.05, 97.15  & 43.13, 72.39	 & 40.68, 73.53  & 77.07, 88.60\\
ViT-B/16-\textcolor{cyan}{\textit{\textbf{F}}} & 50.17, 78.74	& 49.00, 80.29	 & 81.48, 97.15 & 45.78, 71.87 & 43.74, 74.41	 & 74.93, 90.60 \\
ResNet-101 & 37.43, 70.17	&	32.92, 70.28	 & 77.49, 96.72  & 35.74, 64.70 & 31.16, 66.46	 &  71.94, 90.46\\
ResNet-101-\textcolor{cyan}{\textit{\textbf{F}}}	& 41.43, 74.30 &	41.99, 77.85	 &  77.07, 97.15 &  38.65, 68.65 & 38.86, 72.78	 &  71.23, 90.60\\
ResNet-50	& 29.74, 61.17	& 29.79, 66.27	 & 59.97, 92.17 & 28.91, 58.13 & 28.91, 65.02	 & 55.27, 83.19 \\
ResNet-50-\textcolor{cyan}{\textit{\textbf{F}}}	& 32.52, 62.48 &	34.92, 70.21 	 & 61.40, 90.31 & 32.70, 61.52 & 32.67, 66.77	 & 60.97, 84.90 \\
\hline
\end{tabular}
}
\vspace{5mm}
\caption{Benchmarking results on joint object segmentation and few-shot classification in terms of top-1 and top-5 classification accuracy with \emph{non-episodic testing} on the OCID dataset~\cite{suchi2019easylabel}. For CLIP-based models, different image encoder backbones are tested: ResNet~\cite{he2016deep}, EfficientNet~\cite{tan2019efficientnet} style ResNet (RN50x4, RN50x16, RN50x64) and Vision Transformers (ViT-B/16, ViT-B/32, ViT-L/14)~\cite{dosovitskiy2020image}.}
\label{table:segmentation}
\end{table}

\clearpage

\input{supp-figures/real-world-testing}

\section{Lists of Object Classes} 
\label{sec: dataset}

\subsection{Synthetic Data}
\label{sec: syn-data-details}
\input{supp-figures/synthetic}

\subsection{Real-World Data}
\label{sec: real+ocid-data-details}
\input{supp-figures/real+ocid}

\bibliography{example}  